\documentclass[letterpaper, preprint,  10pt]{AAS} 
\usepackage{amsmath, bm, graphicx, hyperref, overcite}
\PaperNumber{25-542}

\usepackage{graphicx}
\usepackage{amsmath}
\usepackage[version=4]{mhchem}
\usepackage{siunitx}
\usepackage{longtable,tabularx}
\usepackage{multirow}
\usepackage{bm}
\usepackage{amsthm}

\usepackage{mathrsfs}
\usepackage{bigstrut}
\usepackage{booktabs}
\usepackage{lineno}
\usepackage{threeparttable}

\usepackage{algorithm}
\usepackage{algorithmic}
\usepackage{amsmath,amssymb}

\usepackage{siunitx}     
\usepackage{adjustbox}   
\usepackage[margin=1in]{geometry} 

\usepackage{amsthm}
\usepackage{color}

\usepackage{xcolor}
\usepackage{hyperref}
\usepackage[utf8]{inputenc}
\usepackage{textcomp}

\usepackage{graphicx}
\usepackage{amsmath}
\usepackage[version=4]{mhchem}
\usepackage{siunitx}
\usepackage{longtable,tabularx}
\usepackage{multirow}
\usepackage{bm}
\usepackage{amsthm}

\usepackage{mathrsfs}
\usepackage{bigstrut}
\usepackage{booktabs}
\usepackage{lineno}
\usepackage{threeparttable}
\usepackage{algorithm}
\usepackage{algorithmic}
\usepackage{amsmath,amssymb}

\usepackage{siunitx}     
\usepackage{adjustbox}   

\usepackage{amsthm}
\usepackage{color}

\usepackage{xcolor}
\usepackage{hyperref}

\newcommand{\symbf}[1]{\bm{#1}}
\newcommand{\increment}{\Delta}
\usepackage{makecell}

\usepackage{algorithm}

\newcolumntype{T}{S[table-format=1.3, round-mode=places, round-precision=3]}
\newcolumntype{I}{S[table-format=3.0, round-mode=places, round-precision=0]}

\begin{document}

\title{Neural Approximators for Low-Thrust Trajectory Transfer Cost and Reachability}

\author{Zhong Zhang\thanks{Postdoctoral Fellow, Department of Aerospace Science and Technology, Politecnico di Milano, 20156 Milan, Italy.}, 
\ and Francesco Topputo\thanks{Professor, Department of Aerospace Science and Technology, Politecnico di Milano, 20156 Milan, Italy; Senior Member AIAA.}}

\maketitle

\begin{abstract}
     In trajectory design, fuel consumption and trajectory reachability are two key performance indicators for low-thrust missions. This paper proposes general-purpose pretrained neural networks to predict these metrics.
  The contributions of this paper are as follows: 
  Firstly, based on the confirmation of the Scaling Law applicable to low-thrust trajectory approximation, the largest dataset is constructed using the proposed homotopy ray method, which aligns with mission-design-oriented data requirements. 
 Secondly, the data are transformed into a self-similar space, enabling the neural network to adapt to arbitrary semi-major axes, inclinations, and central bodies. This extends the applicability beyond existing studies and can generalize across diverse mission scenarios without retraining.
 Thirdly, to the best of our knowledge, this work presents the current most general and accurate low-thrust trajectory approximator, with implesmentations available in C++, Python, and MATLAB. 
  The resulting neural network achieves a relative error of 0.78\% in predicting velocity increments 
   and 0.63\% in minimum transfer time estimation. The models have also been validated on a third-party dataset,  multi-flyby mission design problem, and mission analysis scenario, demonstrating their generalization capability, predictive accuracy, and computational efficiency.
\end{abstract}

\section{Introduction}

Low-thrust electric propulsion has gained significant attention due to its higher specific impulse and improved efficiency over traditional high-thrust chemical propulsion. As a result, it has been widely used in real-world missions such as Deep Space 1 \cite{raymanMissionDesignDeep1999}, Hayabusa \cite{kawaguchiHayabusaItsTechnology2008}, BepiColombo \cite{benkhoffBepiColomboComprehensiveExploration2010}, and Psyche \cite{oh2019development}. 
Despite these advantages, low-thrust propulsion poses significant challenges for mission designers. Unlike high-thrust systems, optimizing low-thrust trajectories involves solving more complex optimal control problems. During preliminary mission design, it is often necessary to evaluate a large number of trajectory candidates under varying mission constraints. As a result, computing millions of low-thrust trajectories becomes a common requirement that can be computationally prohibitive.

 As physicist Lev Landau noted, “the most important part of doing physics is the knowledge of approximation,” a principle that equally applies to this case. To address such computational burden, approximate methods are commonly employed in low-thrust mission design to estimate key performance indicators such as fuel consumption and trajectory feasibility, thereby avoiding the need to solve complex optimal control problems directly.
These approximate methods can then be broadly classified into three categories\cite{zhangGlobalTrajectoryOptimization2024}: analytical approaches, database-driven techniques, and neural network–based models.

Classical analytical techniques—such as Lambert two-impulse solutions~\cite{izzoRevisitingLambertsProblem2015,woollands2015new,russellCompleteLambertSolver2022}, Edelbaum’s formulas and their refinements~\cite{edelbaum1961propulsion}, shape-based trajectory methods~\cite{wallShapeBasedApproachLowThrust2009,wuAnalyticalShapingMethod2022}, and other simplifications of system dynamics~\cite{hennesFastApproximatorsOptimal2016,casalino2014approximate,shen2021simple,gurfil2004analysis}—offer closed-form or semi-analytical approximations. While computationally efficient, these methods often suffer from limited accuracy and generality, for instance, being restricted to specific cases such as near-circular or near-coplanar transfers.

Database-driven methods offer high accuracy and efficiency for well-defined problem classes~\cite{petropoulos_anastassios_2018_1139152,zhangGTOC11Results2023a}; however, their generalization capability is limited, as performance depends heavily on the coverage and data generation method\cite{zhangEfficientLowThrustTrajectory2025}. Furthermore, in many mission scenarios, task parameters cannot be fully specified in advance, often necessitating frequent database updates, which further limits the adaptability of these methods.

In recent years, an increasing number of researchers have turned to machine learning techniques, particularly deep learning, to develop efficient surrogate models for trajectory evaluation. Deep neural networks, as universal function approximators~\cite{parkUniversalApproximationUsing1991}, have been shown capable of capturing complex nonlinear relationships with high accuracy and have been applied this astrodynamics tasks.
Zhu et al. \cite{zhuFastEvaluationLowThrust2019} proposed a learning‑based approach to assess low-thrust transfer feasibility and optimal fuel consumption within a specified time window. Their method integrates a classification network for feasibility with a regression network for fuel cost, and experimental results showed that the networks replicated outcomes of expensive trajectory optimizations.
Li et al. \cite{liDeepNetworksApproximators2020} used neural networks to approximate three optimal transfer costs: minimum time‑of‑flight, minimum fuel consumption, and minimum $\Delta v$ for $J_2$-perturbed multi‑impulse transfers.
More recently, Guo et al. \cite{guoDNNEstimationLowThrust2023} targeted time‑optimal, multi‑leg, multi‑asteroid rendezvous missions by embedding a neural network to predict the minimum time‑of‑flight for transfers between asteroids. To enhance precision on critical trajectories, they emphasized short‑duration transfers—limiting the dataset and applying custom loss functions to improve accuracy.
Acciarini et al. \cite{acciariniComputingLowthrustTransfers2024a} compared neural and analytical methods, emphasizing their advantages and publicly releasing their training dataset to foster open research in neural network applications.
Above studies highlight the effectiveness of neural models as efficient approximators, capable of providing real-time transfer-cost estimates to support complex mission design. Similar deep learning techniques have also been applied successfully in other astrodynamics tasks, such as orbit uncertainty, periodic orbit generation in multi-body dynamics, on-board guidance, and autonomous navigation\cite{harlNeuralNetworkBased2013,wilsonGenerationClassificationCritical2024,federiciImageBasedMetaReinforcementLearning2022,izzoRealTimeGuidanceLowThrust2021,izzoNeuralRepresentationTime2023,pugliattiOnboardStateEstimation2024}.

Despite significant progress in recent years, existing research still faces several limitations. First, validation procedures for neural networks rely on self-generated or closely related datasets. Such validation strategies tend to artificially inflate performance measures due to the shared data distribution between training and test sets. Second, existing models are designed for certain mission configurations, such as fixed specific impulse or a narrow thrust range, with insufficient exploration of their generalizability across varied scenarios. A general model that eliminates the need for data regeneration and retraining when addressing new missions would significantly accelerate application. Third, the lack of publicly available models hinders the reproducibility and verifiability of existing studies, thereby limiting the credibility of neural network-based approaches among mission designers.


This paper aims to develop robust neural network models that can be applied across diverse space mission scenarios without the need for regenerating or retraining, thereby significantly reducing time in practical mission design.
To this end, the contributions of this paper are threefold. 
First, a dataset of 100 million samples is constructed using the proposed homotopy ray method, with its distribution aligned with typical mission design requirements. 
Second, to enhance model generalization, the data is transformed into a self-similar space, enabling the neural network to generalize across varying semi-major axes, inclinations, and central bodies.
Third, 
to the best of the authors’ knowledge, this will be the first publicly available low-thrust trajectory approximator, with implementations provided in C++, Python, and MATLAB. This open access facilitates comparative analysis and encourages broader adoption and trust in neural network methods within the space community.

In addition to strong performance on the internal test set, the proposed approach is further evaluated using a third-party dataset. It is further applied to the multi-asteroid flyby problem from the 4th Global Trajectory Optimization Competition (GTOC4), where it help to yield the best-known solution to date. Finally, the models are benchmarked against a direct optimal control method via pork-chop plot analysis in an asteroid rendezvous scenario.



\section{Data Generation}
\label{sec:DataGeneration}

This section presents a data generation method for producing large-scale, mission-oriented trajectories. An indirect method is first used to solve individual trajectories, followed by an efficient sampling strategy called the Homotopy Ray Method to generate a diverse set of solutions. In practice, trajectory data is biased toward regions of interest, such as those with low fuel consumption or minimal transfer time. The proposed method concentrates on densely sampling trajectories in these preferred regions.

Two types of data are generated for two separate neural networks: one for fuel-optimal problems and the other for time-optimal problems. Note that trajectory reachability can essentially be framed as a time-optimal control problem. If the given transfer time is shorter than the time-optimal value, the trajectory is unreachable; conversely, it is reachable when the transfer time exceeds this value. When the transfer time equals the time-optimal solution, the trajectory lies at the boundary of the reachable set, as shown in~\cite{zhuFastEvaluationLowThrust2019}. Thus, the reachability analysis in~\cite{zhuFastEvaluationLowThrust2019} can be regarded as a special case of time-optimal prediction. Compared to a binary classification of reachable versus unreachable, predicting the minimum transfer time offers richer physical insights. Accordingly, reachability prediction in this work explicitly corresponds to solving the time-optimal control problem.

Data generation was implemented in C++ and executed on a workstation featuring an AMD EPYC 7452 processor (2.6 GHz, 64 cores) and 256 GB of memory.
\subsection{Indirect Method for Optimal Control}
This subsection describes the indirect method used to solve two classical low-thrust optimal control problems. In both cases, Pontryagin's minimum principle is applied by introducing the costate vector to transform the original optimal control problem into a two-point boundary value problem (TPBVP), which is then solved by a shooting method.

\subsubsection{Problem Formulations}

In all the problem formulations, we assume that the maximum thrust $T_{\max}$ and specific impulse $I_{\rm sp}$ are constants (i.e., independent of the distance between spacecraft to the central body). The state vector $\boldsymbol{x}$ represents the spacecraft spatial state. 
 $\boldsymbol{x}_0$ and $\boldsymbol{x}_f$ denote the initial and final state, respectively. 

For the time-optimal problem, the goal is to minimize the transfer time.


Before introducing the indirect method for solving optimal control problems, four practical techniques are used to facilitate the solution process. 
\begin{enumerate}
  \item Instead of using Cartesian coordinates $(\boldsymbol{r}, \boldsymbol{v})$ or classical orbital elements $(a, e, i, \Omega, \omega, f)$, the state vector $\boldsymbol{X}$ is represented using modified equinoctial elements (MEE)\cite{walkerSetModifiedEquinoctial1985}, which are non-singular and well-suited for solving majority of low-thrust trajectory optimization problems\cite{junkinsExplorationAlternativeState2019}.
  \begin{equation}
    \begin{array}{ccc}
        p = a (1 - e^2), & f = e \cos (\omega + \Omega), & g = e \sin (\omega + \Omega), \\
        h = \tan (i/2) \cos \Omega, & k = \tan (i/2) \sin \Omega, & L = \omega + \Omega + \upsilon,
    \end{array}
\end{equation}
where $a$, $e$, $i$, $\Omega$, $\omega$, and $\upsilon$ denote the semi-major axis, eccentricity, inclination, right ascension of the ascending node, argument of perigee, and true anomaly, respectively.
Defining the spacial state vector as $\bm{x} = [p,\,f,\,g,\,h,\,k,\,L]^{\mathrm{T}}$, the dynamics are expressed as
\begin{equation}
    \label{eq:indirect_dynamics}
    \dot{\bm{x}} = \bm{D}(\bm{x}) + \frac{T_{\max}}{m}  \bm{M}(\bm{x})\,\bm{\alpha} u
\end{equation}
where  $\bm{D}$ represents the gravitational dynamics, $\bm{\alpha}$ is the unit vector indicating the thrust direction,
${u}$ is the normalized thrust magnitude, i.e. $\bm{u}=\bm{\alpha}u$ and $u \in [0, 1]$,
 $\bm{M}$ is the transformation matrix linking the control to the MEE rates. 
Detailed matrix expressions for $\bm{M}$ and $\bm{D}$ can be found in Appendix. Readers interested in the various formulations of these expressions may refer to the works of~\cite{izzoRealTimeGuidanceLowThrust2021,junkinsExplorationAlternativeState2019,gaoLowThrustInterplanetaryOrbit2004} for further details.

  \item To address the discontinuity of bang-bang control in fuel-optimal problems, a logarithmic homotopy method is applied to smooth the control profile, thereby improving convergence and solution efficiency~\cite{bertrandNewSmoothingTechniques2002}.
  \item To facilitate the initialization of the costate vector, a normalization technique is employed by introducing a scalar multiplier $\lambda_0$, such that the magnitude of the initial costate vector $\boldsymbol{\lambda}_0$ is set to 1. This does not alter the nature of the optimal control problem, but improves the ability to estimate reasonable initial guesses for the costate values \cite{jiangPracticalTechniquesLowThrust2012a}. Therefore, the fuel-optimal cost function is reformulated as
  \begin{equation}
    \label{eq:indirect_log_homotopy}
    J_{\rm fuel} =       \lambda_0 \int_{t_0}^{t_f} L_{\rm fuel}(\bm{x}, \bm{u})\mathrm{d} t =
      \lambda_0 \int_{t_0}^{t_f} \frac{T_{\max }}{I_{\mathrm{sp}} g_0}  \{u-\varepsilon \ln [u(1-u)]\} \mathrm{d} t
  \end{equation}
with $\epsilon = 1\times 10^{-5}$ in all generation process, which effectively smooths the control profile and maintains the high fidelity~\cite{bertrandNewSmoothingTechniques2002}. The time-optimal cost function is similarly defined as
\begin{equation}
  \label{eq:indirect_log_homotopy_time}
  J_{\rm time} =  
    \lambda_0 \int_{t_0}^{t_f} 1 \mathrm{d} t
\end{equation}

\item In the time-optimal problem, the final states are represented as functions of $t_f$ to enhance feasibility and are enforced as equality constraints $\boldsymbol{x}(t_f) = \boldsymbol{x}_f(t_f)$, in contrast to the fixed terminal states in the fuel-optimal problem $\boldsymbol{x}(t_f) = \boldsymbol{x}_f$.
\end{enumerate}

To summary up, the fuel-optimal and time-optimal  problem can be expressed as
\paragraph{Fuel-Optimal Problem:}
\begin{align}
  \min_{\boldsymbol{u}} \quad & J_{\rm fuel} =  
      \lambda_0 \int_{t_0}^{t_f} \frac{T_{\max }}{I_{\mathrm{sp}} g_0} \left\{ u - \varepsilon \ln \left[u(1-u)\right] \right\} \, \mathrm{d}t, \label{eq:final_fuel_obj} \\[1mm]
  \text{s.t.} \quad 
  &     \dot{\bm{x}}(t)  = \bm{D}(\bm{x}) + \frac{T_{\max}}{m}  \bm{M}(\bm{x})\,\bm{\alpha} u(t) , \label{eq:final_fuel_dyn} \\[1mm]
  & \dot{m}(t) = -\frac{T_{\max} \, {u}(t)}{I_{\mathrm{sp}} g_0}, \label{eq:final_fuel_mass} \\[1mm]
  & \boldsymbol{x}(t_0) = \boldsymbol{x}_0, \quad m(t_0) = m_0, \label{eq:final_fuel_init} \\[1mm]
  & \boldsymbol{x}(t_f) = \boldsymbol{x}_f. \label{eq:final_fuel_term}
\end{align}

\paragraph{Time-Optimal Problem:}
\begin{align}
  \min_{\boldsymbol{u},\, t_f} \quad & J_{\rm time} =  
      \lambda_0 \int_{t_0}^{t_f} 1 \, \mathrm{d}t, \label{eq:final_time_obj} \\[1mm]
  \text{s.t.} \quad 
  &     \dot{\bm{x}}(t) = \bm{D}(\bm{x})+ \frac{T_{\max}}{m}  \bm{M}(\bm{x})\,\bm{\alpha} u(t), \label{eq:final_time_dyn} \\[1mm]
  & \dot{m}(t) = -\frac{T_{\max} \, {u}(t)}{I_{\mathrm{sp}} g_0}, \label{eq:final_time_mass} \\[1mm]
  & \boldsymbol{x}(t_0) = \boldsymbol{x}_0, \quad m(t_0) = m_0, \label{eq:final_time_init} \\[1mm]
  & \boldsymbol{x}(t_f) = \boldsymbol{x}_f(t_f). \label{eq:final_time_term}
\end{align}

\subsubsection{Indirect method and TPBVP Formulation}
The indirect method is employed to solve the above two optimal control problem, which involves transforming the original problem into a two-point boundary value problem (TPBVP) using Pontryagin's minimum principle. 

For the fuel-optimal problem, introducing the costate vector $\bm{\lambda}_x(t), {\lambda}_m(t)$ and scalar multiplier $\lambda_0$, the Hamiltonian is constructed as
\begin{equation}
  H_{\rm fuel} = \bm{\lambda}_x^{\mathrm{T}} \dot{\bm{x}} + \bm{\lambda}_m \dot{m} + \lambda_0  L_{\rm fuel}(\bm{x}, \bm{u}) = 
  \frac{T_{\max}}{m} \bm{\lambda}_x^{\mathrm{T}} \bm{M}\,\bm{\alpha} u + \bm{\lambda}_x^{\mathrm{T}} \bm{D} - \lambda_m \frac{T_{\max}}{I_{\rm sp} g_0}\boldsymbol{\alpha}u + \lambda_0 \frac{T_{\max }}{I_{\mathrm{sp}} g_0}  \{u-\varepsilon \ln [u(1-u)]\}.
\end{equation}

The costate dynamics are then given by
\begin{equation}
  \label{eq:indirect_costate}
\begin{aligned}
    \dot{\bm{\lambda}}_x &= -\frac{\partial H}{\partial \bm{x}} = -  ( \frac{ T_{\max} }{m} \boldsymbol{\lambda}^T \frac{\partial \boldsymbol{M}}{\partial \boldsymbol{x}}  \boldsymbol{\alpha}u +\boldsymbol{\lambda}^T \frac{\partial \boldsymbol{D}}{\partial \boldsymbol{x}} )\\
  \dot{{\lambda}}_m &= -\frac{\partial H}{\partial {m}} = \frac{T_{\max}}{m^2} \bm{\lambda}_x^{\mathrm{T}} \bm{M}\,\bm{\alpha} u.    
\end{aligned}
\end{equation}
where the detailed expressions of partial derivative of $\bm{M}$ and $\bm{D}$ to $\bm{x}$ can be refered to the Appendix.


Now, the control is unknown in costate dynamics Eq.~\eqref{eq:indirect_costate}.
According to Pontryagin's Minimum Principle, the Hamiltonian must be minimized respect to control, which leads to the optimal thrust direction and magnitude:
\begin{equation}
    \label{eq:indirect_optimal_control}
    \bm{\alpha}^* = -\frac{\bm{M}^{\mathrm{T}} \bm{\lambda}_x}{\|\bm{M}^{\mathrm{T}} \bm{\lambda}_x\|}, 
\end{equation}
\begin{equation}
  \label{eq:indirect_optimal_control_u}  
  u^* = \frac{2\epsilon}{\rho+2\epsilon+\sqrt{\rho^2+4\epsilon^2}}.
\end{equation}
where $\rho$ is the switching function and expressed by:

\begin{equation}
\rho = 1 - \frac{I_{\rm sp} g_0 \|\bm{M}^{\mathrm{T}} \bm{\lambda}_x\| }{\lambda_0 m } - \frac{\lambda_m}{\lambda_0}
\end{equation}

For rendezvous problems, the state vector must satisfy the boundary conditions
\begin{equation}
	\bm{x}(t_0) = \bm{x}_0, \quad \bm{x}(t_f) = \bm{x}_f.
\end{equation}

Since the terminal mass is free in the problem, the transversality condition is provided by 
\begin{equation}
  \lambda_m(t_f) = 0.
\end{equation}

In addition, as suggested in \cite{jiangPracticalTechniquesLowThrust2012a}, the augmented costate vector 
\[
\bm{\lambda}(t) \triangleq \begin{bmatrix}  \bm{\lambda}_x(t) \\  \lambda_m(t) \\ \lambda_0 \end{bmatrix}
\]
is normalized at the initial time:
\begin{equation}
    \|\bm{\lambda}(t_0)\| = 1.
\end{equation}

Consequently, the fuel-optimal control problem is converted into the following two-point boundary value problem:
\begin{equation}
    \label{eq:indirect_TPBVP}
    \varPhi_{\rm fuel}\left[\bm{\lambda}(t_0)\right] = 
    \begin{bmatrix}
        \bm{x}(t_f) - \bm{x}_f \\
        \lambda_m(t_f) \\
		\|\bm{\lambda}(t_0)\| - 1
    \end{bmatrix}
    = \bm{0}.
\end{equation}
This TPBVP is solved via a shooting method \cite{MoreJ.J.GarbowB.S.andHillstrom1980}. Once the optimal initial costate $\bm{\lambda}_x(t_0)$ and final time $t_f$ are determined, the state and costate equations \eqref{eq:indirect_dynamics} and \eqref{eq:indirect_costate} can be integrated to obtain the complete trajectory and corresponding control law. Finally, the remaining mass is computed as
\begin{equation}
	m(t_f) = m_0 - \dot{m} \cdot (t_f - t_0).
\end{equation}

For the time optimal problem, only the differences are presented because of majority similar derivations. The Hamiltonian is defined as
\begin{equation}
  H_{\rm time} =  
  \frac{T_{\max}}{m} \bm{\lambda}_x^{\mathrm{T}} \bm{M}\,\bm{\alpha} u + \bm{\lambda}_x^{\mathrm{T}} \bm{D} - \lambda_m \frac{T_{\max}}{I_{\rm sp} g_0}\boldsymbol{\alpha}u + \lambda_0.
\end{equation}

The costate dynamics and the optimal control direction remain the same as in fuel-optimal problem Eqs.(\ref{eq:indirect_costate}-\ref{eq:indirect_optimal_control}), while the normalized magnitude of optimal control keeps the constant to 1 in time-optimal problem.

With reference to the transversality conditions provided in \cite{levineControlSystemsHandbook2018}, the optimal final time $t_f$ is determined by
\begin{equation}
    H_{\rm time}(t_f) - \bm{\lambda}_x(t_f) \cdot \dot{\bm{x}}_f = H_{\rm time}(t_f) - \bm{\lambda}_L(t_f)(\frac{\sqrt{\mu p}}{r^2}) = 0.
\end{equation}

As a result, the time-optimal control problem is converted into the following two-point boundary value problem:
\begin{equation}
  \label{eq:indirect_TPBVP_time}
  \varPhi_{\rm time}\left[\bm{\lambda}(t_0); t_f\right] = 
  \begin{bmatrix}
      \bm{x}(t_f) - \bm{x}_f \\
      \lambda_m(t_f) = 0 \\
      H_{\rm time}(t_f) - \bm{\lambda}_L(t_f)(\frac{\sqrt{\mu p}}{r^2}) \\
  \|\bm{\lambda}(t_0)\| - 1
  \end{bmatrix}
  = \bm{0}.
\end{equation}

\subsection{Homotopy Ray Method for Large-Scale Data Generation}
To tackle the challenge of generating large-scale datasets suitable for trajectory design, the Homotopy Ray Method is proposed. The primary goal is to efficiently generate mission-design-oriented trajectory samples that exhibit both low fuel consumption and close to the boundary of reachable trajectories. This data distribution plays a critical role in numerical trajectory optimization and will be illustrated using a practical example later.


\subsubsection{Scaling Law in Low-Thrust Trajectory Approximation}


The necessity of large-scale datasets is first examined by verifying whether the scaling law applies to low-thrust trajectory estimation. The Scaling Law, originally introduced by OpenAI during research on large language models, reveals that increasing the scale of computation, model size, and dataset can directly lead to performance improvements\cite{kaplanScalingLawsNeural2020}. However, this phenomenon has not been thoroughly explored in the context of low-thrust trajectory optimization. In fact, reference\cite{zhuFastEvaluationLowThrust2019} suggests that dataset size an efficient trade-off point for dataset scaling.

This paper examines the impact of dataset and model size, using fuel-consumption estimation as a representative case. The neural network inputs, outputs, and training methodology are described in detail later.

\begin{figure}[hbt!]
  \centering
  \includegraphics[width=0.58\textwidth]{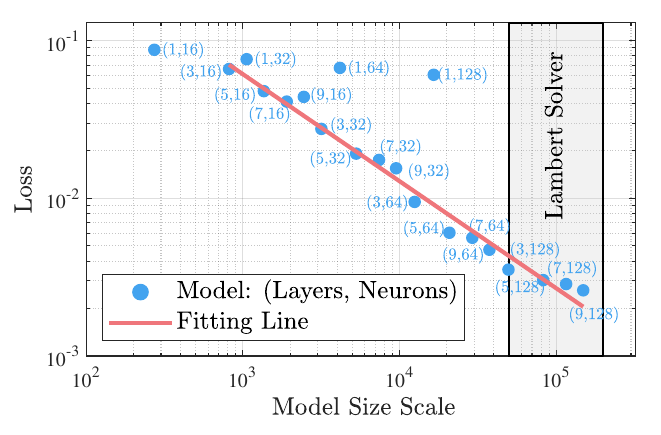}
  \caption{Model size scalling law in low-thrust approximation}
  \label{fig:scalinglaw_model}
\end{figure}

\begin{figure}[hbt!]
  \centering
  \includegraphics[width=0.6\textwidth]{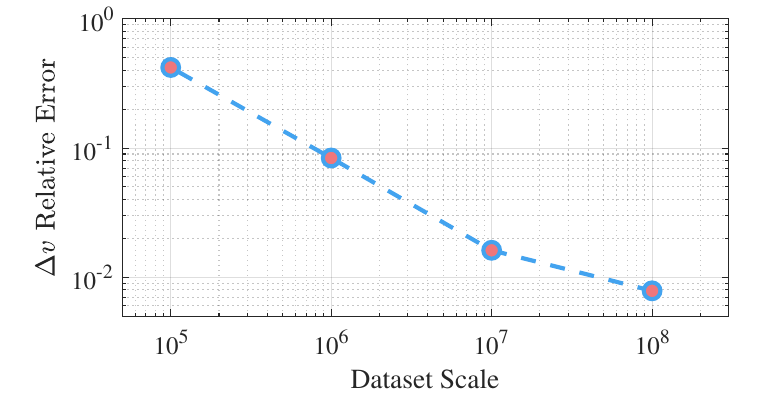}
  \caption{Dataset size scalling law in low-thrust approximation}
  \label{fig:scalinglaw_datasize}
\end{figure}


As shown in Figures \ref{fig:scalinglaw_model} and \ref{fig:scalinglaw_datasize}, both dataset and model size have a significant impact on performance.
Figure~\ref{fig:scalinglaw_model} illustrates how the loss of model decreases as its size increases, where each blue dot represents a neural network configuration defined by the number of layers and neurons. The shaded area indicates the range in which the model's inference time is on the same order of magnitude comparable to that of the Lambert solver\cite{izzoRevisitingLambertsProblem2015}.
Notably, performance exhibits an approximately linear growth on a logarithmic scale, supporting the validity of the Scaling Law in this domain. For each case, multiple model configurations were tested, and the best-performing hyperparameters were selected.
During the experiments, it was also observed that the growth in these two dimensions is independent; increasing the dataset size does not affect the choice of model size, and vice versa. These findings underscore the importance of both dataset and model size in low-thrust trajectory optimization.

However, model size cannot be increased indefinitely, as larger models lead to longer inference times, which becomes a bottleneck when used as a submodule within complex optimization frameworks. The model size is selected to ensure its inference time remains on the same order of magnitude as that of a Lambert solver, as shown in the gray region in Fig.~\ref{fig:scalinglaw_model}, thereby maintaining a balance between predictive efficiency and mission-level computational throughput. 

As a result, an appropriate model size—namely, 9 hidden layers with 128 neurons each—is selected, and the subsequent focus shifts to scaling up the dataset. It is worth noting that increasing dataset size only affects training time, not inference time. Since training is conducted once offline and the model is reused subsequently, training time is generally not a concern for end users.



\subsubsection{Large-Scale Data Generation}
\label{sec:KeplerianNeighborhood}

\begin{figure}[hbt!]
  \centering
  \includegraphics[width=0.5\textwidth]{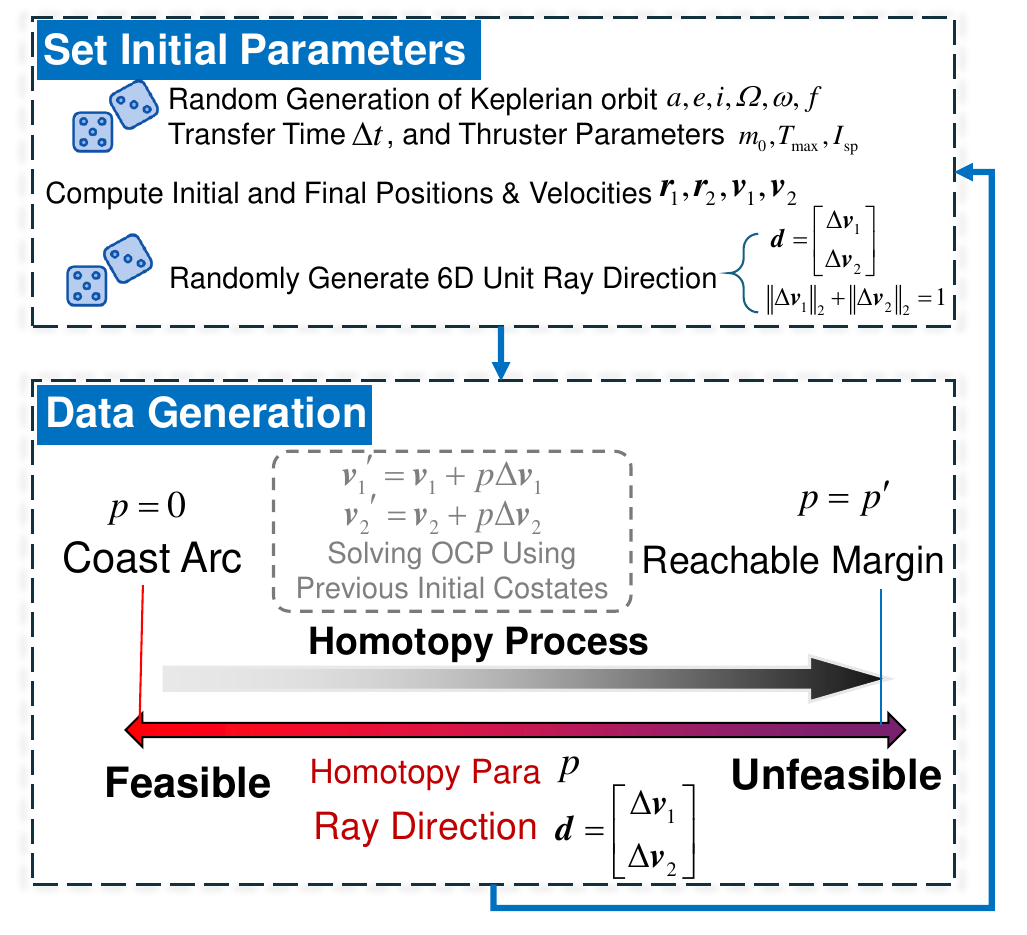}
  \caption{Schematic diagram of the homotopy ray method.}
  \label{fig:homotopy_generate}
\end{figure}

This section aims to expand the dataset size to enhance solution quality and convergence robustness. A two-phase method is proposed: it begins with low-fuel trajectories and gradually deforms them toward near-infeasible solutions, utilizing previously obtained effective initial costates to facilitate faster convergence of the shooting method. The method consists of two primary components, as illustrated in Fig.~\ref{fig:homotopy_generate}: (1) generation of an initial orbit that lies within a Keplerian neighborhood, and (2) homotopic continuation toward the boundary of infeasible trajectories.

\begin{enumerate}
  \item Firstly, random initial guesses are generated through a method termed Keplerian Orbital Neighborhood Sampling. Due to the thrust limitations of low-thrust propulsion systems, arbitrarily assigned initial and final states often lead to infeasible transfers. As a result, randomly sampling initial and terminal states results in considerable computational waste and rarely includes fuel-efficient trajectories.
  An intuitive idea is that, in certain extreme cases, when the initial and final states lie on the same natural Keplerian orbit, the spacecraft can complete the transfer without any propulsion effort, ensuring feasibility regardless of the thrust magnitude.

  Under the two-body assumption, such orbits correspond to classical Keplerian trajectories.During data generation, a Keplerian orbit is randomly sampled with orbital elements $(a, e, i, \Omega, \omega, f)$ and a transfer duration $\increment t$, which is then converted into the corresponding initial and final position-velocity vectors $(\boldsymbol{r_1}, \boldsymbol{v_1}, \boldsymbol{r_2}, \boldsymbol{v_2})$.

  \item   Secondly, a homotopy ray method is developed to generate large-scale, mission-oriented trajectory datasets.
  During the data generation process, the algorithm exploits the concept of homotopy by reusing the initial costate values from previously solved trajectories, and progressively deforms feasible trajectories toward infeasible ones.

  Since Keplerian trajectories are guaranteed to be feasible, the process begins with such a trajectory. A random six-dimensional normalized direction vector is generated as $d = [\increment \symbf{v_1}; \increment \symbf{v_2}]$, and a homotopy parameter $p$ is used to control the deviation of the boundary velocities from the original Keplerian values. That is, in the optimal control formulation, the boundary velocities are set as $\symbf{v}_1' = \symbf{v}_1 + p \increment \symbf{v}_1$ and $\symbf{v}_2' = \symbf{v}_2 + p \increment \symbf{v}_2$. By gradually increasing the homotopy coefficient $p$, a sequence of trajectory data is obtained along the direction defined by $d$, until the resulting trajectory becomes infeasible for both probelms and extreme large value for time-optimal problem. By repeating first and second procedure with different Keplerian orbits and directions $d$, a large amount of high-quality trajectory data can be efficiently generated.
\end{enumerate}




The specific algorithmic implementation details can be found in Algorithm~\ref{alg:HomotopyRay}. For a randomly selected Keplerian orbit, normalized velocity increments $\increment \symbf{v}_1$ and $\increment \symbf{v}_2$ are generated such that their squared norms sum to one. A small initial scaling factor $p$ is set to 10 m/s, and the scaled increments $p\increment \symbf{v}_1$ and $p\increment\symbf{v}_2$ are added to the original velocity vectors $\symbf{v}_1$ and $\symbf{v}_2$.
The initial costate vector $\symbf{\lambda}_0$ is obtained by solving the shooting equation~Eq.~\eqref{eq:indirect_TPBVP} or Eq.~\eqref{eq:indirect_TPBVP_time} with multiple random guesses. The value of $p$ is then gradually increased, reusing the previously obtained costate vector $\symbf{\lambda}_0$ to solve the shooting problem directly until it becomes unsolvable.

\begin{algorithm}[htbp!]
  \caption{Homotopy Ray Method}
  \label{alg:HomotopyRay}
  \begin{algorithmic}[1]
    \REPEAT 
    \STATE Randomly generate initial position $\boldsymbol{r}_1$, velocity $\boldsymbol{v}_1$, terminal position $\boldsymbol{r}_2$, velocity $\boldsymbol{v}_2$, transfer time $\Delta t$, initial mass $m$, specific impulse $I_{\rm sp}$ and thrust $T_{\max}$
    \STATE Randomly generate normalized perturbations $\Delta \boldsymbol{v}_1$ and $\Delta \boldsymbol{v}_2$ such that $\|\Delta \boldsymbol{v}_1\|^2 + \|\Delta \boldsymbol{v}_2\|^2 = 1$
    \STATE Initialize the homotopy parameter $p_{\rm start}$ and solve the initial costate $\boldsymbol{\lambda}_0$ for $p_{\rm start}$
    \STATE Initialize the empty set of solved trajectories $\mathcal{S}_{\rm solved}$
    \REPEAT
        \STATE Compute perturbed velocities:
        $
        \boldsymbol{v}_1' = \boldsymbol{v}_1 + p\,\Delta \boldsymbol{v}_1, \quad \boldsymbol{v}_2' = \boldsymbol{v}_2 + p\,\Delta \boldsymbol{v}_2
        $
        \STATE Solve the TPBVP using the current costate guess $\boldsymbol{\lambda}_0$
        \IF{the TPBVP is solved successfully}
            \STATE Update: $p \gets p + \Delta p$
            \STATE Set $\boldsymbol{\lambda}_0$ to the newly obtained costate for the next iteration
            \STATE Store the trajectory in $\mathcal{S}_{\rm solved}$
        \ELSE
            \STATE Reduce the step size: $\Delta p \gets \Delta p/2$
        \ENDIF
    \UNTIL{the problem becomes unsolvable: $\Delta p$ falls below a pre-defined threshold}
    \STATE Select trajectories from $\mathcal{S}_{\rm solved}$ to move to the Data Sets $\mathcal{S}$
    \UNTIL{sufficient data is generated}
    \STATE Output Data Sets $\mathcal{S}$
  \end{algorithmic}
  \end{algorithm}

Note that the unsolvable trajectory corresponds to different physical meanings under the two optimal control formulations.
For the fuel-optimal problem, the value $p'$ is defined as the boundary of the reachable set.
As discussed at the beginning of this section, the computed time-optimal transfer time $t_f$ equals $\increment t$ in this case.
For the time-optimal problem, the trajectory often corresponds to a large fuel consumption value, since the engine is at full thrust at all times. This may even result in an eccentricity greater than one and cause numerical difficulties. To avoid such singular cases, an early termination condition is added for the time-optimal problem.


Finally, the reason for imposing an equality constraint on the terminal position Eq.~\eqref{eq:final_time_term} in the time-optimal problem, rather than fixing it at a specific point, is clarified. Based on the homotopy ray method, the initial trajectory must be feasible. Fixing the terminal position may lead to an infeasible trajectory. 
This is because, in the time-optimal problem, the engine operates at full thrust continuously. If the initial transfer time exceeds the time required to consume all the propellant, the trajectory will naturally fail to reach the terminal point.
To mitigate this risk, the terminal position equality constraint is formulated to ensure the feasibility of the initial trajectory.

\subsubsection{Data Distribution Characteristics}

\begin{figure}[hbt!]
    \centering
    \includegraphics[width=0.75\linewidth]{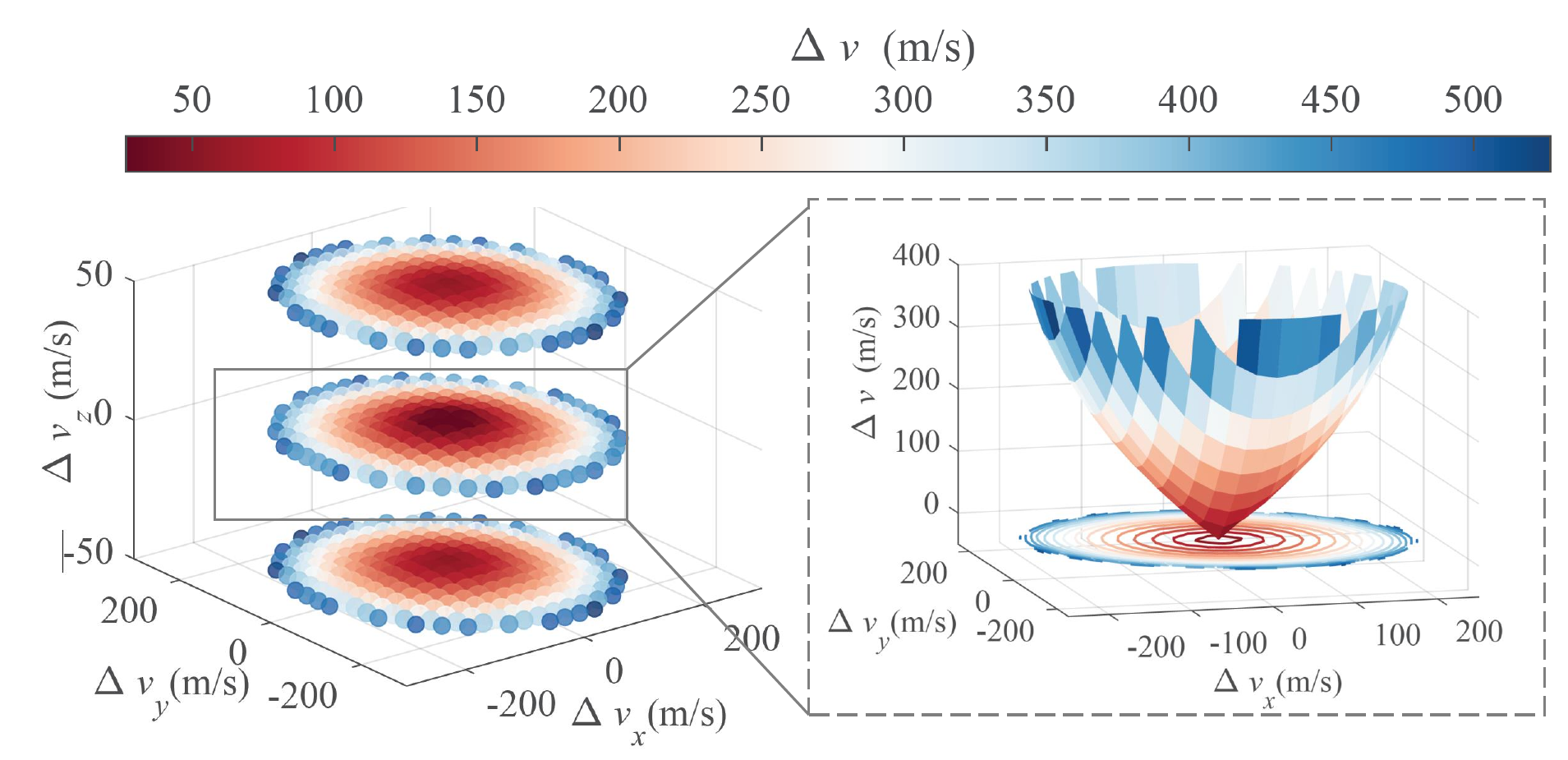}
    \caption{Graph illustrating the relationship between the speed increment and the corresponding variation in departure velocity.}
    \label{fig:Figure_convex_diagram}
\end{figure}

  The homotopy ray method not only accelerates data generation, but also reveals an approximately monotonic relationship between the velocity increment and $\increment\symbf{v}_1$, $\increment\symbf{v}_2$, as shown in Fig.~\ref{fig:Figure_convex_diagram}. Although this is just a specific case, the figure illustrates the variation in fuel consumption $\increment v$ as $\increment\symbf{v}_1$ changes while keeping $\increment\symbf{v}_2 = 0$, demonstrating the monotonic property of the velocity increment.

  This monotonic behavior also motivates the use of $\increment\symbf{v}_1$ and $\increment\symbf{v}_2$ as input features instead of velocity vectors directly, as they better capture the underlying structure of the problem. This is consistent with observations reported in~\cite{acciariniComputingLowthrustTransfers2024a,liDeepNetworksApproximators2020,zhuFastEvaluationLowThrust2019}, which used Lambert solver solutions as the inputs to improve network performance. Details of the input features are provided in the next section.
  
  Another advantage of this approach lies in the adaptively denser sampling of low-fuel-consumption trajectories and unreachable regions, which aligns well with the requirements of practical trajectory optimization. Such a sampling distribution is difficult to achieve through random generation. This advantage will be further demonstrated in the result section.

\section{Input and Output Data Analysis}
\label{sec:input_output_analysis}

This section aims to reduce input dimensionality through lossless transformations, in order to improve model performance. To this end, dimensionality reduction is first introduced in a self-similar space. Next, the model's performance under different input configurations is evaluated. Finally, the input features of the two models along with the intermediate data processing are summarized.

Specifically, the input parameters for Eq.~\eqref{eq:final_fuel_obj} and Eq.~\eqref{eq:final_time_obj} include the following: ${{{\bm{r}}_1},{{\bm{v}}_1},{{\bm{r}}_2},{{\bm{v}}_2},\Delta t,{m_0},{T_{\max }},{I_{{\rm{sp}}}},\mu }$. The total input dimension of this formulation is 17. Note that the gravitational acceleration at sea level $g_0$, which is typically fixed at 9.80665~m/s$^2$, remains unchanged. Minor discrepancies in numerical precision can be compensated by adjusting the specific impulse $I_{\rm{sp}}$, since $g_0$ and $I_{\rm{sp}}$ always appear as a product.

This section aims to leverage physical insights and empirical analysis to identify a minimal set of independent variables. The goal is to determine the optimal combination of input features that preserves model performance.

\subsection{Dimensionality Reduction in Self-Similar Space}
\label{sec:dimensionality_reduction}


This subsection introduces the concept of self-similar mapping, which leverages the inherent rotational invariance and dimensional invariance of the problem. It is important to note that this process involves data preprocessing to reduce the input dimensionality of the neural network, rather than data augmentation. Since these transformations are physically lossless, such dimensionality reduction is both reasonable and justified.

\subsubsection{Rotational Invariance}




A central gravitational field exhibits inherent rotational symmetry, meaning that the system's dynamics depend solely on the relative distance to the center rather than on its absolute orientation in space.
In other words, if both the initial and final positions and their corresponding velocity vectors are rotated by the same angle, their relative geometrical relationships remain unchanged. The physical constraints that govern trajectory optimization also remain unaffected.

Consequently, the optimal solution, whether minimizing the fuel consumption or the transfer time, remains invariant under coordinate system rotation.
This rotational invariance offers flexibility in coordinate selection for trajectory planning and guarantees that the resulting solution is independent of the chosen reference frame.

\begin{figure}[htb!]
  \centering
  \includegraphics[width=0.4\linewidth]{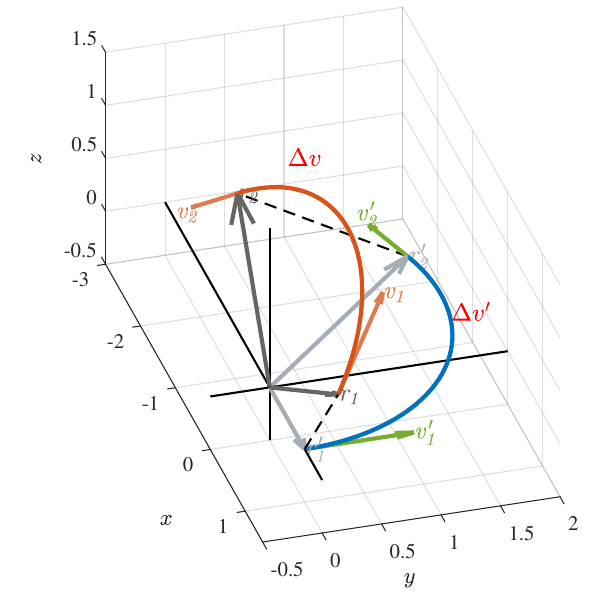}
  \caption{The diagram illustrates the rotational invariance.}
  \label{fig:dimensionless_rotational}
\end{figure}

As illustrated in Fig.~\ref{fig:dimensionless_rotational}, the system dynamic equations remain invariant under coordinate rotation in a dimensionless space.
A rotation applied to the initial and final states $\boldsymbol{r_1}, \boldsymbol{v_1}, \boldsymbol{r_2}, \boldsymbol{v_2}$ yields the transformed states $\boldsymbol{r_1'}, \boldsymbol{v_1'}, \boldsymbol{r_2'}, \boldsymbol{v_2'}$.
Since the dynamic equations are invariant under rotation, the velocity increment of the transformed problem is equal to that of the original one, namely, $\Delta v = \Delta v'$. The same conclusion applies to time-optimal solutions.


There are several strategies for coordinate rotation. In this study, the rotation is performed by aligning the departure position with the x-axis and the arrival position with the xy-plane. This transformation effectively reduces the problem dimensionality by three without any loss of physical fidelity.

Since transfer trajectories in different planes can be expressed in a unified formulation, this approach improves the generality and consistency of model prediction.

The detailed procedure is outlined in Algorithm~\ref{alg:rotation}. The core idea involves rotating the initial position onto the x-axis, followed by a secondary rotation about the x-axis to place the final position within the xy-plane. Among the two possible orientations resulting from the second rotation, the one that yields a positive projection of the initial velocity vector onto the y-axis is selected.

\begin{algorithm}[H]
  \caption{Rotation $\boldsymbol{r_1},\boldsymbol{v_1},\boldsymbol{r_2},\boldsymbol{v_2}$}
  \label{alg:rotation}
  \begin{algorithmic}[1]
  \REQUIRE Initial states: $\boldsymbol{r_1},\boldsymbol{v_1}$; Final states: $\boldsymbol{r_2},\boldsymbol{v_2}$
  \ENSURE Rotated states: $\boldsymbol{r_1'},\boldsymbol{v_1'};\;\boldsymbol{r_2'},\boldsymbol{v_2'}$
  \STATE Compute $\boldsymbol{n} \gets \mathrm{Cross}(\boldsymbol{r_1},\boldsymbol{r_2})$
  \STATE Normalize: $\boldsymbol{n} \gets \boldsymbol{n}/\|\boldsymbol{n}\|$
  \STATE Remove normal component from $\boldsymbol{v_1}$: 
  $
  \boldsymbol{v_1} \gets \boldsymbol{v_1} - \big(\mathrm{Dot}(\boldsymbol{v_1},\boldsymbol{n})\big)\boldsymbol{n}
  $
  \STATE Set target direction: $\boldsymbol{e_x} \gets [1,\,0,\,0]$
  \STATE Compute rotation matrix $R_1 \gets \mathrm{RotationMatrix}(\boldsymbol{r_1},\boldsymbol{e_x})$
  \STATE Obtain intermediate states:
$
  \boldsymbol{r_1}^* \gets R_1\,\boldsymbol{r_1},\quad \boldsymbol{v_1}^* \gets R_1\,\boldsymbol{v_1}, \quad 
  \boldsymbol{r_2}^* \gets R_1\,\boldsymbol{r_2},\quad \boldsymbol{v_2}^* \gets R_1\,\boldsymbol{v_2}
$
  \STATE Compute correction angle: $\phi \gets -\arctan2\big(v_{1z}^*,\,v_{1y}^*\big)$
  \STATE Compute $R_x \gets \mathrm{RotateX}(\phi)$
  \STATE Compose total rotation: $R_{\text{total}} \gets R_x\,R_1$
  \STATE Obtain final states:
$
  \boldsymbol{r_1'} \gets R_{\text{total}}\,\boldsymbol{r_1},\quad \boldsymbol{v_1'} \gets R_{\text{total}}\,\boldsymbol{v_1},\quad
  \boldsymbol{r_2'} \gets R_{\text{total}}\,\boldsymbol{r_2},\quad \boldsymbol{v_2'} \gets R_{\text{total}}\,\boldsymbol{v_2}
$
  \RETURN $\boldsymbol{r_1'},\boldsymbol{v_1'},\boldsymbol{r_2'},\boldsymbol{v_2'}$
  \end{algorithmic}
  \end{algorithm}


\subsubsection{Dimensional Invariance}


This subsection discusses the process of non-dimensionalization, a commonly used technique in trajectory optimization.
A typical approach involves normalizing the length and time units by defining 1 AU and 1 year as the respective reference units~\cite{jiangPracticalTechniquesLowThrust2012a}. Another widely used method scales the gravitational parameter so that the solar gravitational constant is normalized to unity~\cite{izzoRealTimeGuidanceLowThrust2021}.

\begin{figure}[htbp]
  \centering
  \includegraphics[width=0.6\linewidth]{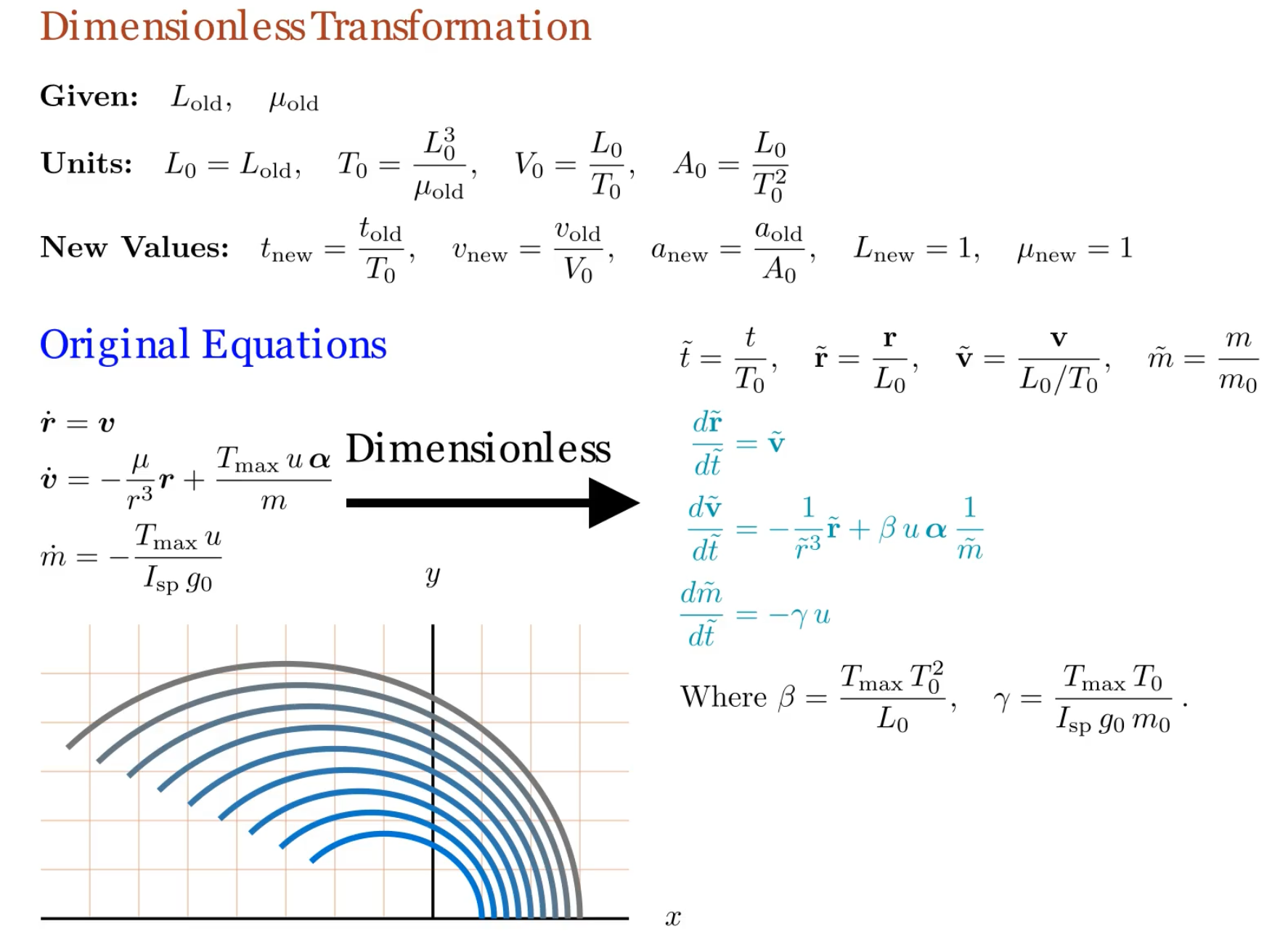}
  \caption{The diagram illustrates the dimensional invariance.}
  \label{fig:dimensionless}
\end{figure}

In this work, a similar normalization scheme is adopted, as shown in Fig.~\ref{fig:dimensionless}. From the perspective of neural network input design, the length unit is defined as the magnitude of the departure position, which normalizes it to unity. Additionally, the gravitational parameter is scaled to unity.
Since both the length and gravitational parameter become fixed after normalization, this further reduces two input dimensions for the neural network.

Consequently, the family of trajectories illustrated in the lower left of Fig.~\ref{fig:dimensionless} can be treated as a single representative case in the normalized space, further enhancing the generalization capability of the proposed model.
Furthermore, since only the ratio between the maximum thrust $T_{\max}$ and the initial mass $m_0$ affects the trajectory, the initial acceleration $a_s = T_{\max} / m_0$ is introduced to further reduce the dimensionality of the corresponding input.
In summary, the original problem involves 17 independent variables. After applying rotational invariance (3 dimensions) and non-dimensionalization (3 dimensions), the independent input dimensionality is reduced to eleven.

\subsection{Comparison of Different Types of Inputs}
\label{sec:comparison_different_inputs}



The previous subsection discussed the minimal dimensionality of input variables. This subsection further presents a performance comparison of models under different input configurations.

\begin{table}[htbp]
  \centering
  \caption{Input Parameters and Descriptions}
    \begin{tabular}{ll}
      \toprule
      Input & Description \\
      \midrule
      $\mathrm{coe}$         & $\text{coe}_1,\ \text{coe}_2$ \\
      $\mathrm{mee}$         & $\text{mee}_1,\ \text{mee}_2$ \\
      $\mathrm{rv}$          & $\boldsymbol{r}_1,\ \boldsymbol{v}_1,\ \boldsymbol{r}_2,\ \boldsymbol{v}_2$ \\
      $\mathrm{rv_{rotate}}$ & $\boldsymbol{r}_1',\ \boldsymbol{v}_1',\ \boldsymbol{r}_2',\ \boldsymbol{v}_2'$ \\
      Ref.~\cite{acciariniComputingLowthrustTransfers2024a} & $\boldsymbol{r}_1-\boldsymbol{r}_2,\ \boldsymbol{v}_1-\boldsymbol{v}_2,\ \Delta\boldsymbol{v}_1,\ \Delta\boldsymbol{v}_2$ \\
      $\mathrm{Lambert_{cart}}$ & $e,\ f,\ \Delta\boldsymbol{v}_1,\ \Delta\boldsymbol{v}_2$ (in $[x,y,z]^\mathrm{T}$) \\
      $\mathrm{Lambert_{sph}}$  & $e,\ f,\ \Delta\boldsymbol{v}_1,\ \Delta\boldsymbol{v}_2$ (in $[r,\theta,\varphi]^\mathrm{T}$) \\
      $t_\mathrm{Lambert}$    & $(\|\Delta\boldsymbol{v}_1\|+\|\Delta\boldsymbol{v}_2\|)/a_s$ \\
      \bottomrule
    \end{tabular}
  \label{tab:input_params}
\end{table}

Several input configurations are defined, as summarized in Table~\ref{tab:input_params}. There are three noteworthy aspects regarding these configurations:
Firstly, the transfer time $\Delta t$, initial acceleration $a_s$, and specific impulse $I_{\rm sp}$ are shared across all input configurations and are thus omitted from the table.
Secondly, all input variables are non-dimensionalized. Variables that become constant after self-similar transformation are excluded from the final input set to avoid confusion. For example, under the $\mathrm{rv_{rotate}}$ configuration, the departure position $\boldsymbol{r}_1'$ is always mapped to a unit vector $[1, 0, 0]^{\rm T}$ and is therefore omitted.
Thirdly, all angular quantities are encoded using the standard $\sin$/$\cos$ formulation to preserve periodicity and facilitate learning.

\begin{table}[htbp]
  \centering
  \caption{$\Delta v$ Prediction Performance Comparison of Different Input Configurations}
    \begin{tabular}{cccccc}
    \toprule
      & Train & Test & Test & $e_{\rm abs}$, & $e_{\rm rel}$, \\
    Input & Loss & Loss & Loss & m/s & (\%) \\
    \midrule
    $\mathrm{coe}$ & 0.0012  & 0.0034  & 0.0032  & 3.76  & 6.21  \\
    $\mathrm{mee}$ & 0.0010  & 0.0030  & 0.0034  & 3.94  & 6.49  \\
    $\mathrm{rv}$ & 0.0096  & 0.0120  & 0.0113  & 13.21  & 27.52  \\
    $\mathrm{rv_{rotate}}$ & 0.0005  & 0.0026  & 0.0024  & 2.83  & 4.71  \\
    Ref.~\cite{acciariniComputingLowthrustTransfers2024a} & 0.0001  & 0.0021  & 0.0019  & 2.26  & 1.30  \\
    $\mathrm{Lambert_{cart}}$ & 0.0002  & 0.0019  & 0.0017  & 2.00  & 3.58  \\
    $\mathrm{Lambert_{sph}}$ & 0.0002  & 0.0020  & 0.0018  & 2.11  & 1.07  \\
    $\mathrm{rv_{rotate}},\mathrm{Lambert_{sph}}$ & 0.0002  & 0.0021  & 0.0019  & 2.19  & 1.18  \\
    \bottomrule
    \end{tabular}%
  \label{tab:comparison_inputs_dv}%
\end{table}%


The performance of different input configurations on two separate networks, one for $\Delta v$ prediction and the other for $\Delta t$ prediction, is summarized in Tables~\ref{tab:comparison_inputs_dv} and~\ref{tab:comparison_inputs_dt}. Several key observations can be made:
\begin{enumerate}
\item Without incorporating Lambert solver information, the $\mathrm{rv\_rotate}$ configuration clearly outperforms other representations, suggesting that dimensionality reduction substantially improves model performance. Among the commonly used orbital element forms, MEE shows the best performance, while the position–velocity representation performs the worst.

\item With Lambert solver information included, performance improves across all configurations, indicating that this information is highly beneficial.

\item Comparing configurations that incorporate Lambert information reveals that models using only independent variables may suffice to achieve competitive performance. Adding redundant information—such as combining $\mathrm{rv_{rotate}}$ with Lambert inputs—offers no further benefit and may even slightly degrade it, possibly due to increased input dimensionality making training more difficult. This indirectly supports the effectiveness of dimensionality reduction in input design.

\item Across all representations, spherical coordinates consistently outperform Cartesian position–velocity representations, possibly due to their stronger physical interpretability.
\end{enumerate}

\begin{table}[htbp]
  \centering
  \caption{$\Delta t$ Prediction Performance Comparison of Different Input Configurations}
    \begin{tabular}{cccccc}
    \toprule
      & Train & Test & Test & $e_{\rm abs}$, & $e_{\rm rel}$, \\
    Input & Loss & Loss & Loss & days & (\%) \\
    \midrule
    $\mathrm{coe}$ & 0.00025  & 0.00124  & 0.00070  & 0.3093  & 0.300  \\
    $\mathrm{mee}$ & 0.00021  & 0.00065  & 0.00056  & 0.2497  & 0.235  \\
    $\mathrm{rv}$ & 0.00341  & 0.00324  & 0.00414  & 1.8325  & 2.046  \\
    $\mathrm{rv_{rotate}}$ & 0.00010  & 0.00043  & 0.00042  & 0.1842  & 0.119  \\
    Ref.~\cite{acciariniComputingLowthrustTransfers2024a} & 0.00006  & 0.00038  & 0.00038  & 0.1682  & 0.100  \\
    $\mathrm{Lambert_{cart}}$ & 0.00006  & 0.00039  & 0.00039  & 0.1719  & 0.102  \\
    $\mathrm{Lambert_{sph}}$ & 0.00008  & 0.00036  & 0.00038  & 0.1674  & 0.100  \\
    $\mathrm{rv_{rotate}},t_\mathrm{Lambert}$ & 0.00006  & 0.00043  & 0.00038  & 0.1670  & 0.107  \\
    $\mathrm{Lambert_{sph}},t_\mathrm{Lambert}$ & 0.00004  & 0.00038  & 0.00034  & 0.1504  & 0.094  \\
    $\mathrm{rv_{rotate}},\mathrm{Lambert_{sph}},t_\mathrm{Lambert}$ & 0.00004  & 0.00038  & 0.00035  & 0.1533  & 0.095  \\
    \bottomrule
    \end{tabular}%
  \label{tab:comparison_inputs_dt}%
\end{table}%



Based on the above findings, the $\mathrm{Lambert_{sph}}$ configuration is selected as the final input configuration. Even though introducing $t_{\mathrm{Lambert}}$ leads to further performance improvement in the time-optimal problem, $\mathrm{Lambert_{sph}}$ is ultimately selected to ensure consistency across both models.

\begin{figure}[hbt!]
  \centering
  \includegraphics[width=0.6\linewidth]{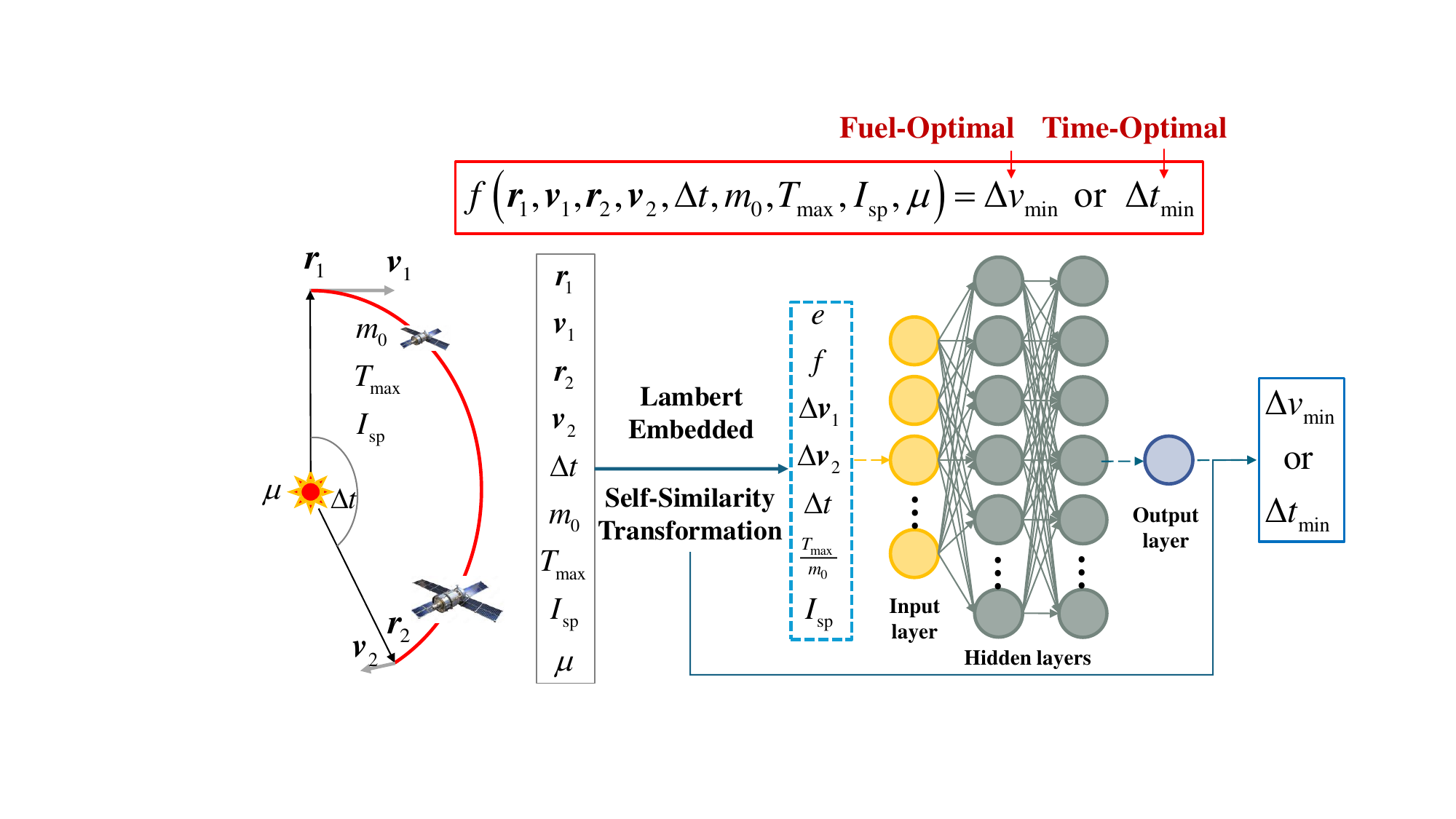}
  \caption{The input and output of the prediction model.}
  \label{fig:input_output}
\end{figure}

Finally, the overall input–output pipeline is illustrated in Fig.~\ref{fig:input_output}. The raw mission parameters are first processed through rotational invariance and non-dimensionalization, reducing the problem dimensionality. Then, a Lambert solver computes the two-impulse transfer characteristics, which are used as inputs to the neural network. The network outputs dimensionless results, which are subsequently rescaled using the original velocity and time units to obtain the final velocity increment and time-optimal transfer duration.

\section{Neural Network Training}
\label{sec:NNTraining}




This section investigates how training-related parameters affect the predictive performance of deep neural networks. The discussion is organized into two parts. The first part evaluates various network architectures, each tested under multiple hyperparameter settings, with only the best result reported per architecture. The second part analyzes the sensitivity of the optimal architecture to individual hyperparameters.

The dataset is split into training, validation, and test sets using a 96\%/2\%/2\% partition. Each model is trained for 10,000 epochs, with the validation set evaluated after each epoch. The model with the best validation performance is retained to prevent overfitting. The retained model is then evaluated on the test set to assess generalization capability. To efficiently evaluate a large number of parameter combinations, these experiments are conducted on a dataset containing 100,000 samples.

All training is performed on a single NVIDIA RTX 4090 GPU (24 GB) with access to 6 CPU cores and 60 GB of memory. The training is implemented using Python 3.12 and PyTorch 2.5.1, with CUDA version 12.4.

\subsection{Neural Network Structure}



\begin{table}[hbt!]
  \centering
  \caption{\textcolor{black}{Ranges of Hyperparameters}}
  \label{table:hyperparameter_ranges}
  \begin{tabular}{cc}
  \toprule
  Hyperparameter & Range \\ \midrule
  $n_{\rm layer}$         & 1, 3, 5, 7, 9  \\ 
  $n_{\rm neuron}$        & 8, 16, 32, 64, 128 \\ 
  $\eta$                  & 0.1, 0.01, 0.001, 0.0001 \\ 
  $wd$ / $\eta$           & 0.1, 0.01   \\ 
  $B$                     & 3200, 6400, 12800, 25600, 51200\\ 
  \bottomrule
  \end{tabular}
\end{table}

The architecture of the network is a critical factor influencing predictive performance. In this subsection, the number of hidden layers ($n_{\rm layer}$) and the number of neurons per hidden layer ($n_{\rm neuron}$) are treated as tunable hyperparameters, with their ranges provided in Table~\ref{table:hyperparameter_ranges}. Additional hyperparameters are introduced in the next subsection.
Each hidden layer uses the ReLU activation function, and a linear activation is applied at the output layer.

The performance of the $\Delta v$ and $\Delta t$ models under different network $n_{\rm layer}$ and $n_{\rm neuron}$ is shown in Tables~\ref{tab:nn_structure_dv} and~\ref{tab:nn_structure_dt}, respectively. Each architecture is fine-tuned over other hyperparameters, and only the best-performing result is reported. The test loss of the $\Delta v$ model has already been presented in Fig.~\ref{fig:scalinglaw_model}. Table~\ref{tab:nn_structure_dt} also presents the performance of the $\Delta t$ model across various architectural configurations. It can be observed that the performance of the $\Delta t$ model continues to improve as the number of layers and neurons increases, satisfying the scaling law as well.

\begin{table}[htbp] 
  \centering
  \caption{ $\Delta v$ Performance for Different Layer and Neuron Configurations}
  \label{tab:nn_structure_dv}
\adjustbox{max width=\textwidth}{
\begin{tabular}{cccccccccccccccc}
\toprule
  & \multicolumn{3}{c}{Neurons = 8} & \multicolumn{3}{c}{Neurons = 16} & \multicolumn{3}{c}{Neurons = 32} & \multicolumn{3}{c}{Neurons = 64} & \multicolumn{3}{c}{Neurons = 128} \\
  \cmidrule{2-16}& Train & Test & Time, & Train & Test & Time, & Train & Test & Time, & Train & Test & Time, & Train & Test & Time, \\
Layers & {Loss} & {Loss} & {s} & {Loss} & {Loss} & {s} & {Loss} & {Loss} & {s} & {Loss} & {Loss} & {s} & {Loss} & {Loss} & {s} \\
\midrule
1 & 0.090  & 0.092  & 156  & 0.085  & 0.087  & 154  & 0.073  & 0.076  & 155  & 0.063  & 0.067  & 154  & 0.057  & 0.061  & 155  \\
3 & 0.073  & 0.072  & 223  & 0.058  & 0.066  & 223  & 0.026  & 0.028  & 225  & {0.006} & {0.009} & {225} & 0.001  & 0.004  & 225  \\
5 & 0.070  & 0.074  & 291  & 0.044  & 0.048  & 292  & 0.017  & 0.019  & 294  & 0.004  & 0.006  & 293  & 0.001  & 0.003  & 295  \\
7 & 0.070  & 0.073  & 360  & 0.038  & 0.041  & 363  & 0.013  & 0.018  & 363  & 0.003  & 0.006  & 362  & 0.001  & 0.003  & 363  \\
9 & 0.070  & 0.070  & 425  & 0.040  & 0.044  & 426  & 0.012  & 0.016  & 431  & 0.002  & 0.005  & 430  & 0.001  & 0.003  & 431  \\
\bottomrule
\end{tabular}%
}
\end{table}

\begin{table}[htbp]
  \centering
  \caption{ $\Delta t$ Performance for Different Layer and Neuron Configurations}
  \label{tab:nn_structure_dt}
\adjustbox{max width=\textwidth}{%
\begin{tabular}{cccccccccccccccc}
\toprule
  & \multicolumn{3}{c}{Neurons = 8} & \multicolumn{3}{c}{Neurons = 16} & \multicolumn{3}{c}{Neurons = 32} & \multicolumn{3}{c}{Neurons = 64} & \multicolumn{3}{c}{Neurons = 128} \\
\cmidrule{2-16}  & Train & Test & Time, & Train & Test & Time, & Train & Test & Time, & Train & Test & Time, & Train & Test & Time, \\
Layers & {Loss} & {Loss} & {s} & {Loss} & {Loss} & {s} & {Loss} & {Loss} & {s} & {Loss} & {Loss} & {s} & {Loss} & {Loss} & {s} \\
\midrule
1 & 0.154  & 0.156  & 157  & 0.137  & 0.146  & 155  & 0.125  & 0.126  & 156  & 0.081  & 0.084  & 155  & 0.0352  & 0.0385  & 156  \\
3 & 0.131  & 0.129  & 224  & 0.091  & 0.096  & 224  & 0.031  & 0.036  & 225  & {0.004 } & {0.005 } & {225} & 0.0007  & 0.0012  & 226  \\
5 & 0.131  & 0.136  & 291  & 0.072  & 0.083  & 290  & 0.013  & 0.017  & 293  & 0.002  & 0.003  & 291  & 0.0005  & 0.0011  & 292  \\
7 & 0.125  & 0.126  & 356  & 0.066  & 0.072  & 355  & 0.009  & 0.011  & 358  & 0.001  & 0.002  & 357  & 0.0006  & 0.0012  & 358  \\
9 & 0.117  & 0.121  & 421  & 0.055  & 0.062  & 422  & 0.007  & 0.010  & 427  & 0.001  & 0.002  & 426  & 0.0004  & 0.0009  & 426  \\
\bottomrule
\end{tabular}%
}
\end{table}

\subsection{Hyperparameter Tuning}
\label{sec:hyperparameter_tuning}



During hyperparameter tuning, the $\Delta v$ and $\Delta t$ models are trained by jointly optimizing three key hyperparameters: the learning rate ($\eta$), the weight decay ratio ($w_d/\eta$), and the batch size ($B$).

The AdamW optimizer, a variant of Adam that incorporates weight decay, is adopted for its robustness in training~\cite{loshchilovDecoupledWeightDecay2018,llugsiComparisonAdamAdaMax2021}. The learning rate $\eta$ and the weight decay ratio $w_d/\eta$ are treated as tunable hyperparameters, with their respective search ranges summarized in Table~\ref{table:hyperparameter_ranges}. A OneCycleLR scheduler is employed to dynamically adjust the learning rate during training~\cite{smithSuperconvergenceVeryFast2019}.

All hyperparameter tuning experiments are conducted using the same network architecture, consisting of 9 hidden layers with 128 neurons each. The test results are reported in Tables~\ref{tab:hyperparameter_dv} and~\ref{tab:hyperparameter_dt}. Given the extensive number of tested cases, only the best-performing settings (highlighted in bold) and corresponding ablation results, where a single hyperparameter is varied while the others are fixed at their optimal values, are shown for clarity.


\begin{table}[htb!]
  \centering
  \caption{Hyperparameter tuning for $\Delta v$}
  \adjustbox{max width=0.45\textwidth}{
    \begin{tabular}{cccccc}
    \toprule
      &   &   & Train & Test & Time, \\
    $\eta$ & $\text{wd}/\eta$ & $B$ & Loss & Loss & s \\
    \midrule
    0.1 & 0.01 & 3200 & 0.0219  & 0.0226  & 424  \\
    0.001 & 0.01 & 3200 & 0.0140  & 0.0160  & 424  \\
    0.0001 & 0.01 & 3200 & 0.0223  & 0.0247  & 421  \\
    \midrule
    \textbf{0.01} & \textbf{0.01} & \textbf{3200} & 0.0062  & 0.0079  & 424  \\
    0.01 & 0.1 & 3200 & 0.0060  & 0.0082  & 424  \\
    \midrule
    0.01 & 0.01 & 6400 & 0.0067  & 0.0092  & 221  \\
    0.01 & 0.01 & 12800 & 0.0081  & 0.0103  & 115  \\
    0.01 & 0.01 & 25600 & 0.0090  & 0.0125  & 62  \\
    0.01 & 0.01 & 51200 & 0.0103  & 0.0127  & 37  \\
    \bottomrule
    \end{tabular}%
    }
  \label{tab:hyperparameter_dv}%
\end{table}%

\begin{table}[htb!]
  \centering
  \caption{Hyperparameter tuning for $\Delta t$}
  \adjustbox{max width=0.45\textwidth}{
    \begin{tabular}{cccccc}
    \toprule
      &   &   & Train & Test & Time, \\
    $\eta$ & $\text{wd}/\eta$ & $B$ & Loss & Loss & s \\
    \midrule
    0.1 & 0.1 & 3200 & 0.0194  & 0.0212  & 426  \\
    0.001 & 0.1 & 3200 & 0.0092  & 0.0114  & 425  \\
    0.0001 & 0.1 & 3200 & 0.0153  & 0.0183  & 427  \\
    \midrule
    \textbf{0.01} & \textbf{0.1} & \textbf{3200} & 0.0033  & 0.0044  & 426  \\
    0.01 & 0.01 & 3200 & 0.0033  & 0.0045  & 426  \\
    \midrule
    0.01 & 0.1 & 6400 & 0.0035  & 0.0045  & 223  \\
    0.01 & 0.1 & 12800 & 0.0043  & 0.0053  & 115  \\
    0.01 & 0.1 & 25600 & 0.0051  & 0.0068  & 62  \\
    0.01 & 0.1 & 51200 & 0.0063  & 0.0080  & 37  \\
    \bottomrule
    \end{tabular}%
  }
  \label{tab:hyperparameter_dt}%
\end{table}%

The results indicate that the learning rate $\eta$ has the greatest impact on model performance, while the influence of the weight decay ratio $w_d/\eta$ is comparatively minor. Due to the low input dimensionality and moderate network size, GPU parallelization enables efficient full-batch processing. As a result, training time scales inversely with the batch size $B$, and smaller batches generally lead to better performance. This introduces a trade-off between computational efficiency and generalization when training on large datasets. In this study, a batch size of $B=6400$ is adopted to accommodate training on datasets as large as 100 million samples.

Finally, these results further demonstrate that careful hyperparameter selection remains essential. Poor choices, such as an excessively large learning rate, can still cause orders-of-magnitude degradation in model performance.

\section{Results} 
\label{sec:results}



This section presents the final training results of the proposed model. It further evaluates the model through three complementary tasks: validation on third-party datasets, application to the GTOC4 multi-flyby asteroid trajectory design problem, and porkchop plot generation for asteroid rendezvous analysis.

The objective is to demonstrate the model’s generalization ability, applicability to real mission design, and computational efficiency. Validation on third-party datasets assesses the model’s generalization under varying data distributions and verifies the effectiveness of the proposed data generation methodology. The GTOC4 multi-flyby design task serves as a benchmark to evaluate the model’s integration within trajectory optimization, highlighting its accuracy in sequential mission planning. Finally, the model is applied to porkchop plot generation for asteroid rendezvous, enabling direct comparison in computational efficiency and usability against traditional optimal control methods.

\subsection{Training Result}
\label{sec:training_result}




The final model is trained on a dataset containing 100 million samples, where generating each 1-million-sample dataset takes approximately three days. The training hardware is the same, and training on the full 100-million-sample dataset also takes around three days to complete.

\begin{table}[htbp]
  \centering
  \caption{$\Delta v$ Model Performance Compared to Existing Models}
  \begin{threeparttable}     
    \begin{tabular}{lcccc}
    \toprule
      & \multicolumn{4}{c}{ $\Delta v$ Model} \\
\cmidrule{2-5}    Parameter & Zhu~\cite{zhuFastEvaluationLowThrust2019} & Li~\cite{liDeepNetworksApproximators2020} & Acciarini~\cite{acciariniComputingLowthrustTransfers2024a} & This paper \\
    \midrule
    $a$, AU & $2.0\sim3.0$ & $2.0\sim3.5$ & $0.9\sim4.0$ & any \\
    $e$ & $0\sim0.4$ & $0\sim0.1$ & $0\sim0.48$ & $0\sim1.0$ \\ 
    $i$, deg & $0\sim20$ & $0\sim10$ & $0\sim30$ & any \\
    $\Delta t,n$ & $0\sim0.48$ & $0\sim0.3$ & $0\sim12.65$ & $0\sim0.99$ \\
    $a_s$ (min), m/s$^2$ & $1.5\times10^{-4}$ & $1.5\times10^{-4}$ & $7.5\times10^{-5}$ & $2.5\times10^{-6}$\tnote{[a]} \\
    $a_s$ (max), m/s$^2$ & $3.8\times10^{-4}$ & $3.0\times10^{-4}$ & $8.6\times10^{-4}$ & $1.2\times10^{-2}$\tnote{[a]} \\
    $I_{\rm sp}$, s & 3000 & 3000 & 4000 & $700\sim9000$\tnote{[a]} \\
    \midrule
    $\Delta v$ MAE, m/s & / & / & 96.16 & 3.38 \\
    $\Delta v$ MRE, \% & / & / & 2.98\% & 0.78\% \\
    mf MAE, kg & / & / & / & 0.42 \\
    mf MRE, \% & 0.37\% & 0.50\% & / & 0.014\% \\
    \bottomrule
    \end{tabular}%
            \begin{tablenotes}    
        \footnotesize             
        \item[a] The value is calculated at 1 AU.      
      \end{tablenotes}
    \end{threeparttable}
  \label{tab:dv_model_final_performance}%
\end{table}%

To assess the applicability and performance of the $\Delta v$ model, we compare our results against those reported in existing studies, as summarized in Table~\ref{tab:dv_model_final_performance}. As shown, our model outperforms previous approaches in terms of both mean absolute error (MAE) and mean relative error (MRE), whether evaluated by $\Delta v$ or the final mass $m_f$. Specifically, the model achieves an average relative error of 0.78\% for $\Delta v$. It should be noted that all comparative results are derived from the respective published test sets, which may overestimate performance due to the likely similarity between training and testing data distributions in those works. 

More significantly, Table~\ref{tab:dv_model_final_performance} also highlights the broad applicability of our model. Unlike prior methods that impose constraints on orbital elements—such as semi-major axis ($a$), eccentricity ($e$), or inclination ($i$)—our model is general-purpose and supports arbitrary celestial bodies. Furthermore, the propulsion parameters ($a_s$, $I_{\rm sp}$) span a wide operational range, covering most practical propulsion configurations. This versatility makes the model suitable for a wide range of mission scenarios, offering substantial convenience compared to earlier models that are often tailored to specific tasks. Mission designers and planetary scientists can directly apply our model without the need for retraining or regenerating datasets. Nevertheless, it is important to note that the current model supports only single-revolution transfers; extending it to multi-revolution cases remains a key direction for future work.

\begin{table}[htbp]
  \centering
  \caption{$\Delta t$ Model Performance}
\begin{threeparttable}
    \begin{tabular}{lc}
    \toprule
       Parameter & $\Delta t$ Model \\
    \midrule
    $a$, AU & any \\
    $e$ & $0\sim1.0$ \\ 
    $i$, deg & any \\
    $\Delta t,n$ & $0\sim0.99$ \\
    $a_s$ (min), m/s$^2$ & $2.5\times10^{-6}$\tnote{[a]} \\
    $a_s$ (max), m/s$^2$ & $1.2\times10^{-2}$\tnote{[a]} \\
    $I_{\rm sp}$, s & $700\sim9000$\tnote{[a]} \\
    \midrule
    $\Delta t\min$ MAE, days & 2.56 \\
    $\Delta t\min$ MRE, \%   & 0.63\% \\
    \bottomrule
    \end{tabular}%
    \begin{tablenotes}    
      \footnotesize             
      \item[a] The value is calculated at 1 AU.      
    \end{tablenotes}
  \end{threeparttable}
  \label{tab:dt_model_final_performance}%
\end{table}%

For the $\Delta t$ model, the evaluation results are reported in Table~\ref{tab:dt_model_final_performance}. Due to the use of equality constraints in the time-optimal control formulation, only our model’s results are presented. The model achieves an average relative error of 0.63\%, demonstrating strong capability in assessing transfer feasibility.

\subsection{Third-Party Dataset Validation}
\label{sec:thirdparty}


This subsection discusses the evaluation of the model’s generalization capability. As discussed in the previous subsection, evaluating the model using its own test set inevitably leads to overestimation, because the test and training sets share the same data distribution. Therefore, a third-party dataset was used for validation. The dataset, provided by \cite{acciariniComputingLowthrustTransfers2024a}, contains one million samples and is currently the only publicly available dataset in this field, representing a significant contribution to the community. The authors also released their model’s prediction results on this dataset, enabling direct comparison. The performance of our model was evaluated on the same dataset and compared with their results. The same evaluation metrics, MAE and MRE, were employed.

\begin{figure}[hbt!]
  \centering
  \includegraphics[width=0.6\linewidth]{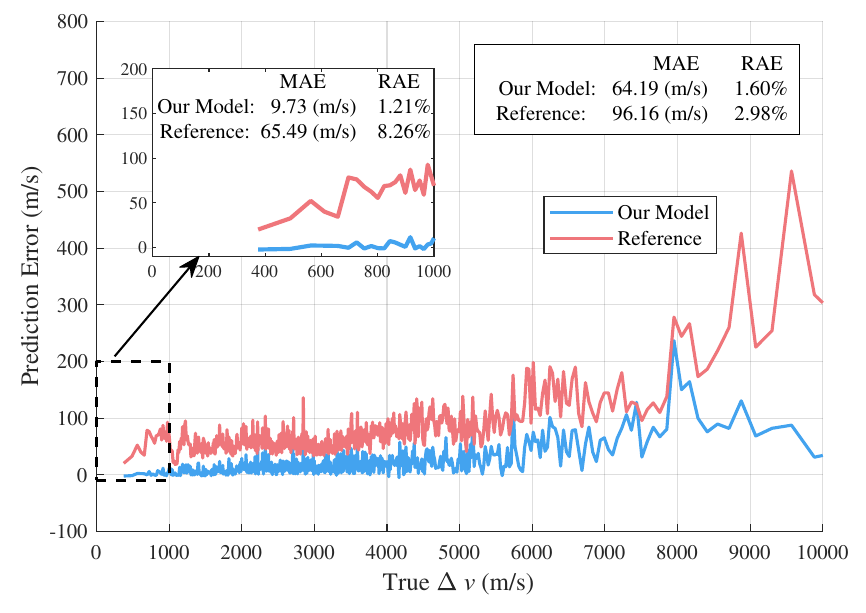}
  \caption{Validation of the third-party dataset.}
  \label{fig:3rdparty}
\end{figure}

As shown in Fig.~\ref{fig:3rdparty}, the model’s performance across various $\Delta v$ was evaluated using a moving average window method. Specifically, the ground-truth $\Delta v$ values were first sorted in ascending order, and the sorted dataset was then divided into consecutive windows, each containing 1000 samples. For each window, the average of the $\Delta v$ values and the corresponding MAE were computed. 
It can be observed that the dataset from \cite{acciariniComputingLowthrustTransfers2024a} does not contain samples with $\Delta v$ less than 200~m/s. This is likely attributed to their data generation method, which relies on randomly selecting initial and final positions and velocities, making low-fuel-consumption samples rare. As a result, their model performs better for $\Delta v$ values above 1000~m/s compared to lower values.

In contrast, the proposed model performs well in this regime, which underscores the importance of the mission design-oriented data generation approach described in the previous section, and demonstrates the effectiveness of the homotopy ray-based data generation method. Moreover, the proposed model achieves better overall performance than that of \cite{acciariniComputingLowthrustTransfers2024a}, even when evaluated on their dataset. This demonstrates the strong generalization ability of the proposed model and further suggests the potential of the scaling law.

\subsection{Multi-Flyby Asteroid Mission Design}
\label{sec:multi-flyby}



This section evaluates the proposed model on a trajectory optimization task. Specifically, the GTOC4 problem is used as a benchmark. This problem involves multiple asteroid flybys, and detailed descriptions can be found in \cite{Grigoriev2013}.
For comparative analysis, two sets of results are considered. One is derived from the state-of-the-art solution to the GTOC4 problem, proposed by the University of Jena. This trajectory includes flybys of 49 asteroids over a ten-year mission, culminating in a final rendezvous. The proposed model predicts the required velocity increments using segment-wise flyby data extracted from this mission plan. This scenario is referred to as Result 1.
Result 2 is generated through neural network-based approximation, which optimizes the timing and relative velocity of each flyby. This results in a more fuel-efficient trajectory, demonstrating the effectiveness of the neural approximators. Since the optimization procedure is beyond the scope of this study, its details are omitted here.

\begin{figure}[hbt!]
  \centering
  \includegraphics[width=0.8\linewidth]{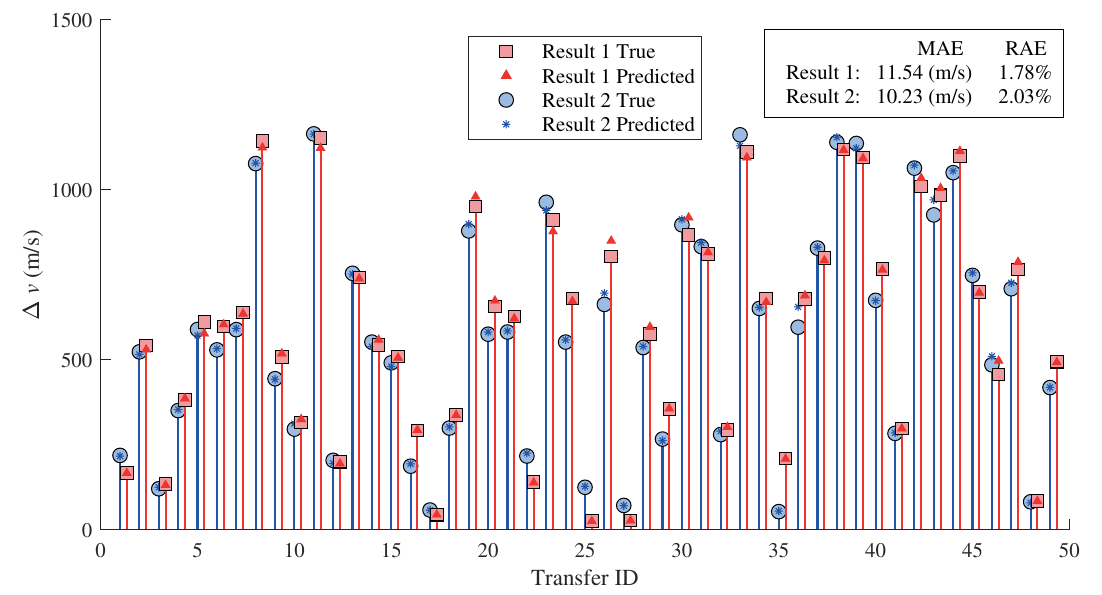}
  \caption{Validate in GTOC4 problem results.}
  \label{fig:GTOC4}
\end{figure}

As shown in Fig.~\ref{fig:GTOC4}, the model demonstrates strong predictive performance on the GTOC4 task. This highlights the effectiveness and generalization capability of the proposed model. In the optimal GTOC4 trajectory, some flyby segments require less than 100~m/s of velocity increment. This underscores the importance of including such low-thrust trajectories in the training dataset.

\subsection{Mission Analysis}
\label{sec:porkchop}

\begin{figure}[hbt!]
  \centering
  \includegraphics[width=1.0\linewidth]{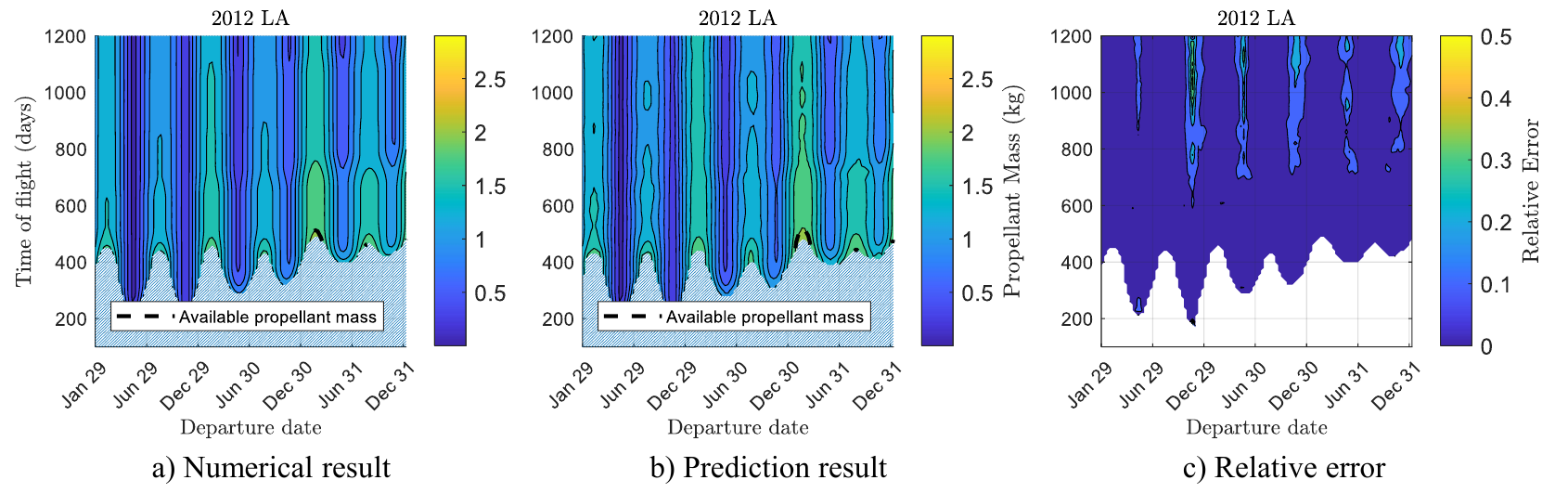}
  \caption{Porkchop plot for the 2012 LA Asteroid.}
  \label{fig:porkchop_2012LA}
\end{figure}

This subsection introduces how the model can be applied to a practical mission analysis scenario. 
The benchmark mission involves a cube satellite launched from Earth to rendezvous with an asteroid, equipped with a low-thrust engine. The launch vehicle provides a maximum departure velocity relative to Earth of 4 km/s. In this paper, asteroid 2012 LA is chosen to show performance, as shown in Figure~\ref{fig:porkchop_2012LA}.




The plot demonstrates that the neural network can approximate the structure of the porkchop plots with high accuracy (typically within 10\%), thereby validating the effectiveness of the proposed model.


\section{Conclusion}
\label{sec:conclusion}

This paper presents a neural network-based approach to efficiently estimate fuel consumption and transfer reachability for low-thrust trajectories. The proposed models extend the existing applicability of low-thrust trajectory approximation, thereby offering adaptability across a variety of mission design scenarios without necessitating retraining for each specific case.
This improvement is driven by two insights. First, at the data level, an approximate scaling law for low-thrust trajectory approximation is validated, supporting the generation of an extensive dataset of over 100 million trajectory samples. Second, at the astrodynamics level, the inherent symmetries in the fuel- and time-optimal control problem are exploited to reduce the dimensionality of the corresponding dynamical system.
The generalizability, accuracy, and computational efficiency of the proposed neural network models are conclusively demonstrated through validation using a third-party dataset, the GTOC4 mission design problem, and a pork-chop mission analysis scenario.

\bibliographystyle{AAS_publication}   
\bibliography{references}   

\end{document}